\def\paperID{30679}
\def\confName{CVPR}
\def\confYear{2026}
\def\paperTitle{A Cross-view Fusion Framework for Robust 6-DoF Grasp Pose Estimation}

\def\authorBlock{
    Kangjian Zhu$^{1}$,
    Haobo Jiang$^{*2}$,
    Jianjun Qian$^{1}$,
    Jin Xie$^{*3}$ \\[10pt]
    $^{1}$Nanjing University of Science and Technology,
    $^{2}$Nanyang Technological University, \\
    $^{3}$Nanjing University \\
    {\tt\small \{zkangjian, csjqian\}@njust.edu.cn, \tt\small haobo.jiang@ntu.edu.sg, csjxie@nju.edu.cn}
}

\newif\ifreview 
\newif\ifarxiv \newcommand{\arxiv}{\arxivtrue}
\newif\ifcamera 
\newif\ifrebuttal 

\arxiv

\pdfoutput=1
\documentclass[10pt,twocolumn,letterpaper]{article}
\ifreview \usepackage[review]{cvpr} \fi
\ifarxiv \usepackage[pagenumbers]{cvpr} \fi
\ifrebuttal \usepackage[rebuttal]{cvpr} \fi
\ifcamera \usepackage{cvpr} \fi

\usepackage{graphicx}	
\usepackage{amsmath}	
\usepackage{amssymb}	
\usepackage{booktabs}
\usepackage{times}
\usepackage{microtype}
\usepackage{epsfig}
\usepackage{caption}
\usepackage{float}
\usepackage{placeins}
\usepackage{color, colortbl}
\usepackage{stfloats}
\usepackage{enumitem}
\usepackage{tabularx}
\usepackage{xstring}
\usepackage{multirow}
\usepackage{xspace}
\usepackage{url}
\usepackage{subcaption}
\usepackage{xcolor}
\usepackage[hang,flushmargin]{footmisc}
\usepackage{pifont}
\usepackage{stfloats}

\ifcamera \usepackage[accsupp]{axessibility} \fi

% ...

\ifarxiv  \fi

\newcommand{\R}[1]{{%
    \textbf{%
        \ifstrequal{#1}{1}{\textcolor{red}{R#1}}{%
        \ifstrequal{#1}{2}{\textcolor{blue}{R#1}}{%
        \ifstrequal{#1}{3}{\textcolor{magenta}{R#1}}{%
        \ifstrequal{#1}{4}{\textcolor{teal}{R#1}}{%
                           \textcolor{cyan}{R#1}%
        }}}}%
    }%
}}

\definecolor{match}{HTML}{30C0B4}
\definecolor{non-match}{HTML}{E54CC9}
\definecolor{MyDarkBlue}{rgb}{0,0.08,1}
\definecolor{MyDarkGreen}{rgb}{0.02,0.6,0.02}
\definecolor{MyDarkRed}{rgb}{0.8,0.02,0.02}  

\usepackage{xr-hyper}

\makeatletter
\newcommand*{\addFileDependency}[1]{
  \typeout{(#1)}
  \@addtofilelist{#1}
  \IfFileExists{#1}{}{\typeout{No file #1.}}
}

\makeatother

\definecolor{cvprblue}{rgb}{0.21,0.49,0.74}
\usepackage[pagebackref,breaklinks,colorlinks,allcolors=cvprblue]{hyperref}
\usepackage[capitalize]{cleveref}
\crefname{section}{Sec.}{Secs.}
\crefname{table}{Table}{Tables}
\crefname{figure}{Fig.}{Figs.}

\ifarxiv \crefname{appendix}{App.}{Apps.}
\else \crefname{appendix}{Suppl.}{Suppls.} \fi

\begin{document}
\title{\paperTitle}
\author{\authorBlock}
\maketitle
\let\thefootnote\relax
\footnotetext{
\hspace*{1em}$^*$Corresponding authors 
}
\begin{abstract}
In this paper, we propose a cross-view fusion framework that enhances the robustness of 6-DoF grasp pose estimation in corner views.
Our framework alleviates occlusion by incorporating an auxiliary view and avoids the time-consuming, task-agnostic multi-view reconstruction through a post-fusion strategy.
To enhance cross-view fusion, we propose a self-supervised contrastive learning strategy that leverages cross-view associations to regularize point cloud features.
In brief, a cross-view point pair is considered a match if the two points correspond to the same 3D location, and a non-match if they represent distinct grasp directions.
The learning strategy significantly enhances the spatial consistency and direction distinctiveness of point features, thereby facilitating cross-view fusion and improving estimation robustness.
Furthermore, we propose a cross-view-aligned cylinder integration module to fuse grasp-relevant geometry into a comprehensive representation.
Specifically, the module first aligns the cross-view points and features according to their similarity to enhance the robustness against noise.
Subsequently, these points are registered into the cylindrical coordinate frame, emphasizing the rotation-symmetric geometry which is important for grasping.
Finally, local self-attention and seed cross-attention layers are alternately employed, respectively enabling interactions within single views and across views, which supports fine-grained representation of grasp-relevant geometry.
Our framework achieves strong performance on the GraspNet-1Billion benchmark and in real-world applications. 
Code is available at \emph{\url{https://github.com/KJZhuAutomatic/Cross-view-Grasp}}.
\end{abstract}

\section{Introduction}
\label{sec:intro}

\begin{figure}[ht]
\centering
\includegraphics[width=\linewidth]{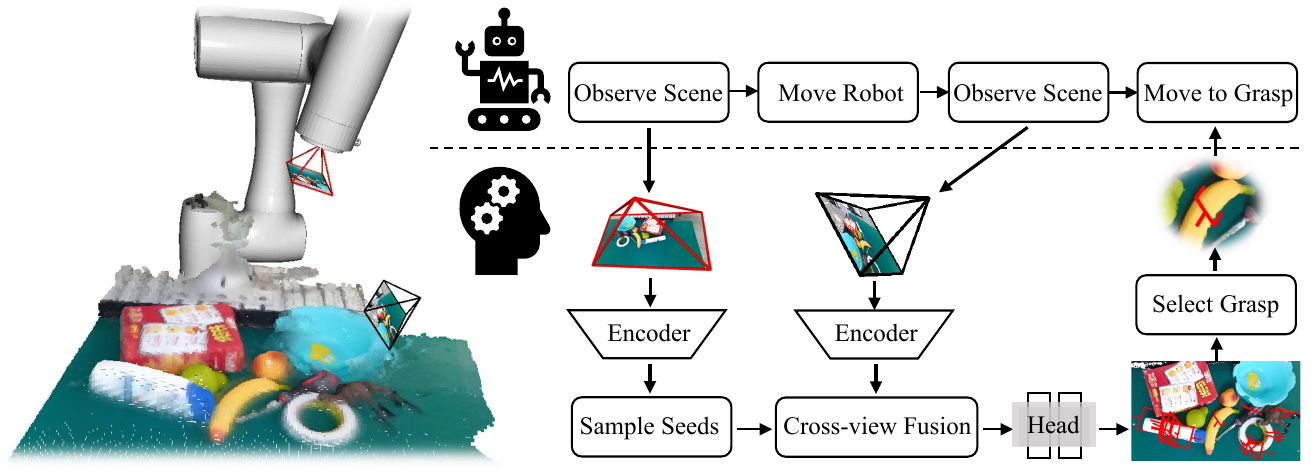}
\caption{ \textbf{Grasp Pipeline with Cross-view Fusion.}
In a robot grasping scenario, a wrist-mounted camera can capture the scene from multiple viewpoints by moving the robot to different joint configurations.
%
The proposed framework fuses cross-view observations to enhance the estimation robustness, while enabling a time-efficient grasp pipeline which requires only an auxiliary-view observation.
}
\vspace{-12pt}
\label{fig:fig1}
\end{figure}

Grasp pose estimation is a fundamental task in many robotic applications, including manipulation \cite{OMG, demonstrate_once}, rearrangement \cite{structdiffusion, IFOR}, and embodied AI \cite{sofar}.
%
Traditional methods predominantly rely on CNN-based detectors to localize graspable rectangles within 2D image. However, these methods are inherently constrained to planar projections and offer only 3-DoF grasp configurations~\cite{planner-grasp-1, planner-grasp-2, planner-grasp-3}. To address these limitations, recent state-of-the-art methods~\cite{graspnet, graspness, graspnet-Subsequent-economic, grasp-reconstruction-2} have shifted toward predicting 6-DoF poses directly from scene point clouds derived from single-view RGB-D observations. 

Despite their efficacy, single-view methods suffer from inherent self-occlusion, which leads to the loss of critical geometric information, a problem particularly pronounced in ``corner-view'' scenarios where the object's structure is partially masked. While robotic mobility allows for capturing the scene from multiple perspectives (\emph{e.g.}, via a wrist-mounted camera), effectively fusing these cross-view observations remains a challenge. In particular, existing multi-view pipelines often resort to ``pre-fusion'' strategies, which first reconstruct a complete 3D scene before performing grasp detection~\cite{Volumetric-Grasping-Network, Volumetric-Grasping-2}. Such strategies, however, are often computationally intensive and may suffer from geometric degradation during the task-agnostic reconstruction process.


In this paper, we propose a high-efficiency cross-view fusion framework for 6-DoF grasp pose estimation. Diverging from the conventional pre-fusion paradigm, we adopt a novel ``post-fusion'' strategy that incorporates an auxiliary view to specifically enhance the robustness of corner-view grasping. 
By performing fusion only within grasp-relevant regions, our framework maintains high-resolution point representations and circumvents the geometric detail loss inherent in pre-fusion full-scene reconstruction.


To effectively bridge the gap between disparate views, we introduce a self-supervised contrastive learning strategy, which regularizes point features by exploiting cross-view associations. Specifically, we define ``matches'' as points corresponding to the same 3D spatial location across views, while ``non-matches'' represent points on the same object but associated with substantially different grasp approach directions. This contrastive objective serves two vital purposes: \textbf{(i)} it enforces spatial consistency in the 3D feature space, preventing corner-view features from drifting out of the training distribution; and \textbf{(ii)} it enhances direction discriminability, enabling the network to distinguish between similar local geometries that require distinct grasp orientations.


Furthermore, we design a cross-view-aligned cylinder integration module to aggregate features within the local grasp-relevant region. Inspired by advanced point cloud registration techniques~\cite{Zero-shot-Registration, jiang2021sampling, jiang2023robust}, we first establish fine-grained correspondences within a cylindrical region to align cross-view features based on coordinate and semantic similarity. This alignment not only identifies overlapping geometries but also mitigates noise from depth sensors and robotic ego-motion through feature averaging. After alignment, we transform the points into a cylindrical coordinate frame to explicitly emphasize rotational symmetry, thereby relieving the network from the burden of inferring orientation from raw Euclidean coordinates. Finally, we encode the point coordinates and process their features using a series of alternating local self-attention and seed cross-attention layers, drawing inspiration from the multi-frame transformer design in VGGT~\cite{vggt} to reduce computational overhead.
Self-attention is used to extract local associations within each single view and its aligned regions, while cross-attention integrates the cross-view context into the seed feature.
This module supports fine-grained point geometry and explicit cross-view interaction, thereby enabling a comprehensive representation to facilitate grasp estimation.

Our main contributions are summarized as follows:
\begin{itemize}[noitemsep, leftmargin=*]
    \item We propose a cross-view fusion framework, which alleviates the occlusion issue in corner views by incorporating an auxiliary view within a time-efficient robotic grasping pipeline, achieving robust 6-DoF grasp pose estimation.
    \item We propose a self-supervised contrastive learning strategy that leverages cross-view associations to regularize point features to be spatially consistent and direction-discriminative, facilitating cross-view fusion and corner-view robustness.
    \item We design a cross-view-aligned cylinder integration module that enables fine-grained interaction of grasp-relevant context while emphasizing rotational symmetry, resulting in a comprehensive representation for accurate estimation.
\end{itemize}
\section{Related Work}
\label{sec:related}

\subsection{Grasp Detection Methods}
%
%
Traditional grasp detection methods \cite{planner-grasp-1, planner-grasp-2, planner-grasp-3} leverage CNNs to detect graspable rectangles in RGB-D images, with the resulting grasp poses constrained to a 2D plane.
%
Benefiting from the advancement of 3D network architectures \cite{PointNet, pointnet-pp, pointcnn, MinkowskiEngine} and their successful applications in 3D detection \cite{zhu2024spgroup3d, Zhu_2025_CVPR}, pose estimation and point cloud registration \cite{jiang2023center, jiang2023se3, jiang2025generative}, recent research has increasingly focused on predicting 6-DoF grasp poses from 3D representations.
%
Some prior work focuses on single-object 6-DoF grasping \cite{object-grasp-1, object-grasp-2, ndf}, which may limit its applicability in broader robotic scenarios.
Another part of the work focuses on predicting 6-DoF grasping in multi-object cluttered scenes \cite{Contact-GraspNet, song2020grasping}, which has long been limited by the lack of real large-scale datasets, until the GraspNet-1Billion benchmark \cite{graspnet} was introduced.
Subsequent works improve the accuracy and generalization by (1) conducting in-depth analysis of point graspability~\cite{graspness}, 
(2) focusing on scale imbalance problem~\cite{graspnet-Subsequent-imbalance}, 
(3) proposing architectures for feature extraction~\cite{graspnet-Subsequent-architecture}, 
(4) utilizing an economic supervision paradigm~\cite{graspnet-Subsequent-economic}.
However, in multi-object cluttered scenes, a single-view observation can only partially represent the scene, leading to the loss of geometric information in corner views, which reduces the robustness of grasp detection.
\subsection{Geometry Enhancement for 6-DoF Grasp}
\noindent
\textbf{Joint Grasping Estimation and Scene Reconstruction.} 
To enable the network to understand the scene geometry from a single-view observation, some studies propose predicting 6-DoF grasping while simultaneously reconstructing the scene \cite{grasp-reconstruction-1, grasp-reconstruction-2, grasp-reconstruction-3}, which can then be used to post-process and refine grasp detection.
While effective, this approach requires scene reconstruction data labels and increases the runtime overhead due to the post-processing step.
In practice, the robot can easily observe the scene from multiple views, using a wrist-mounted camera to construct the grasping scene.
\newline
\textbf{Multi-view Pipelines for Grasping and Manipulation.} 
%
Recently, multi-view observations have been used to enhance scene geometry \cite{jiang2025fuser}, providing improved representations for downstream robotic grasping and manipulation tasks.
VGN \cite{Volumetric-Grasping-Network} and GeneGN \cite{Volumetric-Grasping-2} rely on complete scene representation, which is reconstructed from multiple observations taken by a wrist-mounted camera. 
This leads to significant time consumption for scene reconstruction before the robot can perform effective actions.
%
RVT-v1 and v2 \cite{rvt, rvt2} use multiple calibrated external cameras to reconstruct the scene, typically relying on camera calibration methods such as \cite{MonoSE3}, which increases hardware costs and reduces the flexibility of the working environment.
In this work, we enhance the geometry and grasp robustness by an auxiliary view, without the requirement of complete scene. 

\section{Method}
\label{sec:method}

\begin{figure*}[tp]
    \centering
    \includegraphics[width=\linewidth]{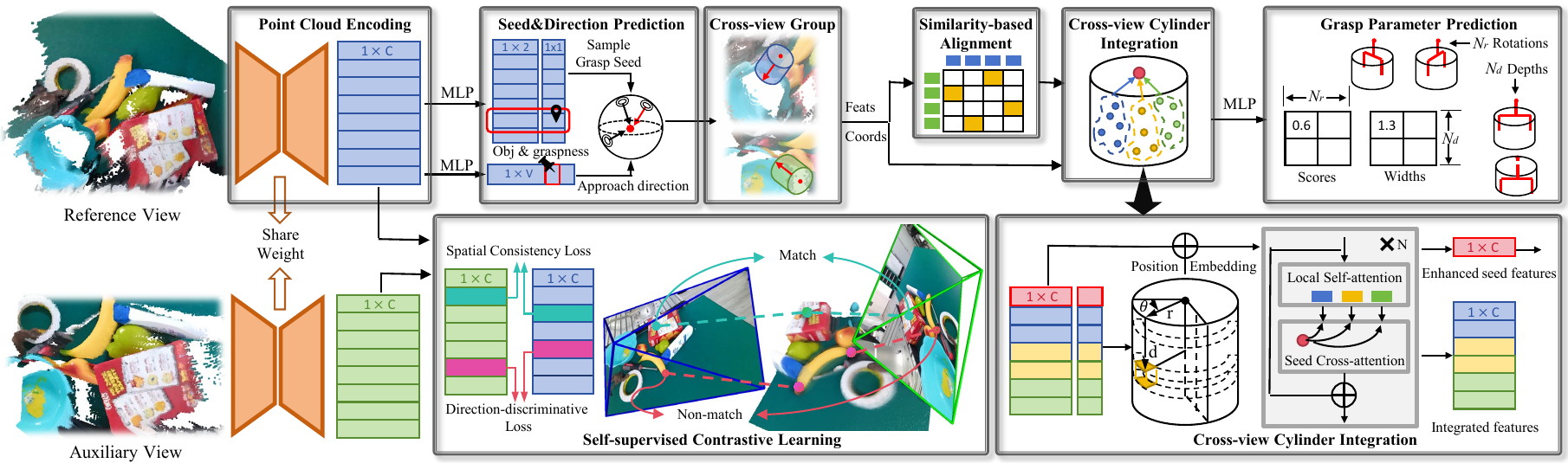}
    \caption{\textbf{Cross-view Grasp Estimation Pipeline}. 
    Given reference and auxiliary views, we first encode point-wise features for both point clouds. On the reference view, objectness and graspability are predicted to sample grasp seeds with approaching directions, followed by neighborhood grouping in both views. Cross-view neighboring points are then matched by feature similarity, embedded in a cylindrical coordinate frame, and processed by alternating attention modules to enrich the seed features with grasp-relevant cross-view context. The enhanced seed features are finally used to predict grasp scores and grasp widths. Moreover, self-supervised contrastive learning leverages cross-view correspondences to regularize point features, improving spatial consistency and directional discriminability for cross-view fusion.
    }
    \vspace{-0.2in}
    \label{fig:fig2}
\end{figure*}

In this section, we present our cross-view fusion framework, which incorporates an auxiliary view to enhance geometric representation and grasp robustness.
%
We first introduce the formulation of the cross-view 6-DoF grasp pose estimation in Sec.~\ref{sec:method-1}.
%
Second, we introduce the grasp estimation framework, where a module integrates robust grasp-relevant representations by aligning and processing cross-view contexts within a cylindrical space, as described in Sec.~\ref{sec:method-2}.
%
Finally, to learn spatially consistent and direction-discriminative point features, we formulate an overall training objective that combines a self-supervised contrastive loss for discovering cross-view associations with a grasp supervision loss, as described in Sec.~\ref{sec:method-3}.
%
%
Our framework is outlined in Fig.~\ref{fig:fig2}.
\subsection{Problem Formulation}
\label{sec:method-1}
%
%
%
%
%

Unlike conventional single-view grasping methods, which often suffer from geometric ambiguity and incomplete observations due to occlusions, we consider a cross-view grasping setting with asymmetric roles across views. Specifically, the reference view (\textit{ref}) serves as the primary view in which grasp candidates are defined and predicted, while the auxiliary view (\textit{aux}) provides complementary geometric cues to enhance grasp reasoning. Their relative transformation can be obtained from the robot's forward kinematics.

For each RGB-D observation, the depth image is back-projected into a colored 3D point cloud, yielding spatial coordinates and corresponding RGB values:
\begin{equation}
\begin{aligned}
    &\mathcal{P}^{ref}=\{p_i^{ref}\mid p_i^{ref}\in\mathbb{R}^3,\ i=1,\dots,N\},\\
    &\mathcal{C}^{ref}=\{c_i^{ref}\mid c_i^{ref}\in\mathbb{R}^3,\ i=1,\dots,N\},
\end{aligned}
\label{eq:input}
\end{equation}
where $N$ denotes the number of points. The auxiliary-view point cloud and color set, $(\mathcal{P}^{aux}, \mathcal{C}^{aux})$, are defined analogously. 
Our goal is to predict a set of valid grasps with respect to the reference view, denoted as $\mathcal{G}=\{\mathbf{g}_s\}_{s=1}^S$, where $S$ is the number of grasp candidates. Each grasp is parameterized as $\mathbf{g}=\{\mathbf{R}, \mathbf{t}, w\}$, where $\mathbf{R}\in SO(3)$ represents the grasp orientation, $\mathbf{t}\in\mathbb{R}^3$ the translation, and $w\in\mathbb{R}$ the gripper width. 
Following GraspNet~\cite{graspnet}, grasp prediction is formulated based on seed points sampled from the reference point cloud. For each grasp seed point $p_s\in\mathcal{P}^{ref}$, the model predicts an approaching direction $\mathcal{A}\in\mathbb{R}^3$, an in-plane rotation angle $r\in\mathbb{R}$, an approaching depth $d\in\mathbb{R}$, and a gripper width $w\in\mathbb{R}$:
\begin{equation}
    \mathcal{A}\in\mathbb{R}^3,\quad r\in\mathbb{R},\quad d\in\mathbb{R},\quad w\in\mathbb{R}.
\end{equation}
The final grasp pose is constructed accordingly: the rotation $\mathbf{R}$ is determined by the approaching direction $\mathcal{A}$ and the in-plane rotation $r$, while the translation $\mathbf{t}$ is obtained by offsetting the seed point $p_s$ along $\mathcal{A}$ with depth $d$.

\subsection{Cross-view Grasp Estimation Framework}
\label{sec:method-2}
\subsubsection{Point Cloud Encoding}
We utilize a ResUNet14~\cite{MinkowskiEngine} backbone with sparse 3D convolutions to extract point-wise geometric features $\mathcal{F}=\{f_i \mid f_i \in \mathbb{R}^{C},i=1,\dots,N\}$ for each view point cloud, as formulated below:
\begin{equation}
\begin{aligned}
    \mathcal{F}^{ref}&=\mathrm{ResUNet}(\mathcal{P}^{ref}, \mathcal{C}^{ref}), \\
    \mathcal{F}^{aux}&=\mathrm{ResUNet}(\mathcal{P}^{aux}, \mathcal{C}^{aux}),    
\end{aligned}
\end{equation}
where $C$ denotes the dimension of point feature.
%
%
ResUNet14 efficiently processes irregular and large-scale point sets via sparse tensor representations, with a U-shaped residual design enabling multi-scale feature learning through downsampling, upsampling, and skip connections.
Despite using ResUNet14 in our implementation, the framework remains compatible with other point-based backbones, such as PointNet~\cite{PointNet} and PointCNN~\cite{pointcnn}.
\subsubsection{Grasp Seed Sampling and Approach Prediction} 

Following GSNet~\cite{graspness}, we apply a Multi-Layer Perceptron (MLP) to each point feature $f_i^{ref}$ from the reference view to predict a 3-dimensional vector independently. 
The first two elements denote the binary classification probabilities $\mathbf{c}_i\in\mathbb{R}^{2}$, indicating whether the point belongs to an object, while the third element encodes the graspable quality score $q_i\in\mathbb{R}$. Points with a score below a predefined threshold are filtered out. Subsequently, we apply Farthest Point Sampling (FPS) to the remaining points in $\mathcal{P}^{ref}$ to obtain a set of grasp seed points and corresponding features $\{(p_s, f_s)\}_{s=1}^S$.  For each grasp seed, a set of predefined approaching directions $ \{\mathcal{A}_i\in\mathbb{R}^{3}|i=1,\dots, N_\mathcal{A}\}$ is evaluated by predicting a quality score for each approaching direction using a shared MLP: $\mathbf{q}^{a}_s=\mathrm{MLP}(f_s)\in\mathbb{R}^{N_\mathcal{A}}$.  During training, the approaching direction is sampled from a categorical distribution based on these scores, whereas during inference, the direction with the highest score is deterministically selected. 
Notably, this grasp seed sampling operation is performed exclusively on the reference view, and in Sec.~\ref{Cross-view-fusion}, we further leverage additional geometric cues from the auxiliary view to complement and enhance the seed features.

\subsubsection{Cross-view-aligned Cylinder Integration}
\label{Cross-view-fusion}
To predict accurate and robust grasps at the seed position $p_s$, we enhance the seed feature $f_s$ by integrating cross-view observation features, forming a comprehensive representation of local grasp-relevant geometry.
\newline
\textbf{Cross-view Cylinder Grouping.}
We group and sample $K$ neighbors and obtain coordinate-feature pairs in a fixed-scale cylindrical region, 
whose position and orientation are determined by the grasp seed's coordinate $p_s$ and approaching direction $\mathcal{A}$. 
%
%
For the auxiliary view, we project its coordinates and features into an aligned coordinate frame via the relative transformation between views.
%
As such, we obtain cross-view-aligned coordinate-feature pairs
$\{(p_k^{ref}, f_k^{ref})\}_{k=1}^K$ 
and 
$\{(p_k^{aux}, f_k^{aux})\}_{k=1}^K$.
%
\newline
\textbf{Similarity-based Alignment.}
In real-world scenarios, depth sensor noise introduces point cloud distortions, and the transformation-based alignment further accumulates robot motion errors, leading to increased noise and degraded geometric representation quality.
%
Following point cloud registration principles, we leverage point coordinates and spatially consistent features to establish cross-view correspondences and compute their averaged coordinates and features to form a robust geometric representation.
%
Specifically, the similarity matrix $ \mathbf{S}\in\mathbb{R}^{K\times K}$ measures similarity through a weighted combination of the coordinate difference and the dot product of normalized features as:
\begin{equation}
\mathbf{S}_{ij}
= -\|p_i^{ref} - p_j^{aux}\|_2^2 
+ \lambda_{\text{feat}} \,
\left\langle 
\frac{f_i^{ref}}{\|f_i^{ref}\|_2}, 
\frac{f_j^{aux}}{\|f_j^{aux}\|_2}
\right\rangle.
\label{eq:similarity}
\end{equation}
We identify the most similar cross-view correspondences and construct aligned coordinate–feature pairs ${(p_k^{ovl}, f_k^{ovl})}_{k=1}^K$ by averaging matched pairs, where the superscript $ovl$ denotes the overlap between the reference and auxiliary views.
\newline
\textbf{Cylinder Coordinate Embedding.}
The above three coordinate-feature pairs $\{(p_k^{ovl}, f_k^{ovl})\}_{k=1}^K$ , $\{(p_k^{ref}, f_k^{ref})\}_{k=1}^K$  and $\{(p_k^{aux}, f_k^{aux})\}_{k=1}^K$ combine cross-view information to represent the seed local geometry, mitigating the risk of missing crucial context from single-view observations.
To avoid the loss of original geometric information in the feature, we encode the 3D coordinate using a learnable MLP-based positional encoder before the processing of multi-head attention.
Unlike directly encoding the coordinates, we first register all the coordinate-feature pairs in the cylindrical coordinate frame, as shown in Fig.~\ref{fig:fig2}.
%
Specifically, for coordinate $p=(x,y,z)$, we convert it by:
\begin{equation}
\theta = \mathrm{atan2}(z, y), \quad
r = \sqrt{y^2 + z^2}, \quad
d = x.
\end{equation}
Notably, the cylindrical coordinate $p' = (\theta, r, d)$ explicitly correspond to the grasp parameters — in-plane rotation, grasp width, and grasp depth, respectively — thereby eliminating the need for the network to infer grasp parameters from Euclidean coordinates.
More importantly, this representation highlights the rotational symmetry of the point cloud, facilitating the extraction and processing of grasp-relevant geometry.
For all coordinate–feature pairs in $\{(p_k^{ovl}, f_k^{ovl})\}_{k=1}^K$, $\{(p_k^{ref}, f_k^{ref})\}_{k=1}^K$, and $\{(p_k^{aux}, f_k^{aux})\}_{k=1}^K$, the coordinate $p=(x,y,z)$ is transformed into the cylindrical coordinate system as $p' = (\theta, r, d)$, and the point features are updated via positional encoding as $\hat{f} = f + \mathrm{Emb}_{\text{pos}}(p')$.
\newline
\noindent
\textbf{Cross-view Context Integration.}
These updated features 
\begin{equation}
\mathcal{F}_K^{ovl} = \{\hat{f}_k^{ovl}\}_{k=1}^K,\mathcal{F}_K^{ref} = \{\hat{f}_k^{ref}\}_{k=1}^K,\mathcal{F}_K^{aux} = \{\hat{f}_k^{aux}\}_{k=1}^K
\end{equation}
are then processed by several attention layers to integrate cross-view contextual information into the grasp seed feature $\hat{f}_s$.
%
Notably, to preserve fine-grained point geometry, the feature sequence is designed to be long in order to retain the original geometric information, which, however, introduces considerable computational overhead in the attention layers.
Inspired by the multi-frame transformer design in VGGT~\cite{vggt}, the features are fed into a series of alternating local self-attention (SA) and seed cross-attention (CA) layers, as formulated below:
\begin{equation}\small
\begin{aligned}
    \hat{\mathcal{F}}_{K}^{ovl}=\text{SA}({\mathcal{F}}_{K}^{ovl}),\ & \hat{\mathcal{F}}_{K}^{ref}=\text{SA}({\mathcal{F}}_{K}^{ref}),\   \hat{\mathcal{F}}_{K}^{aux}=\text{SA}({\mathcal{F}}_{K}^{aux}),\\
    \tilde{f}_s = &\text{CA}(\hat{f}_s, \hat{\mathcal{F}}_{K}^{ovl}, \hat{\mathcal{F}}_{K}^{ref}, \hat{\mathcal{F}}_{K}^{aux}).
\end{aligned}
\end{equation}
The self-attention layer is used to extract local geometric structures from each single view and their overlapping regions, while the cross-attention layer integrates the global context into the seed feature, as shown in Fig.~\ref{fig:fig2}.
The alternating interaction mechanism achieves better performance than the global self-attention layer while significantly reducing computational overhead.
The enhanced grasp seed feature $\tilde{f}_s$ integrates the original single-view feature $f_s$ with cross-view information, resulting in a representation that is significantly enriched in geometric context and facilitates more robust prediction of grasp operation parameters.
\subsubsection{Prediction of Grasp Parameters} 
Based on the predicted approaching direction $\mathcal{A}\in\mathbb{R}^{3}$ centered at the seed coordinate $p_s\in\mathbb{R}^{3}$, the remaining grasp operation parameters include the in-plane rotation $r\in\mathbb{R}$, the approaching depth $d\in\mathbb{R}$, and the gripper width $w\in\mathbb{R}$.
%
Similar to the prediction of the approaching direction, we evaluate a set of predefined rotation–depth pairs by predicting the grasp quality scores. 
%
Specifically, taking the enhanced grasp seed feature $\tilde{f}_s\in\mathbb{R}^C$ as input, an MLP predicts two tensors of shape $N_r\times N_d$, where $N_r$ and $N_d$ denote the number of in-plane rotations and approaching depths, respectively.
%
One of the tensors $\mathbf{w}_s\in\mathbb{R}^{N_r\times N_d}$ represents the predicted gripper widths corresponding to each rotation–depth pair, while the other encodes the grasp quality scores $\mathbf{q}^{g}_s\in\mathbb{R}^{N_r\times N_d}$ associated with these operation parameters.
\subsection{Loss Function}
\label{sec:method-3}
\textbf{Supervised Loss.}
The loss consists of two parts: point-wise supervision for identifying graspable object points, and grasp seed-wise supervision for predicting grasp configurations.
First, we apply point-wise supervision, including object classification $\mathbf{c}_i\in\mathbb{R}^{2}$ and grasp quality regression $q_i\in\mathbb{R}$:
\begin{equation}
    \mathcal{L}_{point} = \mathrm{CrossEntropy}(\mathbf{c}_i, \mathbf{c}_{i,gt})+\lambda_g\mathrm{L}_2(q_i, q_{i,gt}).
\end{equation}
Second, grasp seed-wise supervision is applied to approaching-direction quality scores
$\mathbf{q}^{a}_s\in\mathbb{R}^{N_\mathcal{A}}$, gripper widths
$\mathbf{w}_s\in\mathbb{R}^{N_r\times N_d}$, and grasp quality scores
$\mathbf{q}^{g}_s\in\mathbb{R}^{N_r\times N_d}$ over predefined in-plane rotations and approaching depths:
\begin{equation}
\begin{aligned}
    \mathcal{L}_{seed} = & \lambda_\mathcal{A}\mathrm{L}_2(\mathbf{q}^{a}_s, \mathbf{q}^{a}_{s,gt}) + \\
    & \lambda_w\mathrm{L}_2(\mathbf{w}_s, \mathbf{w}_{s,gt}) +\lambda_\mathcal{G}\mathrm{L}_2(\mathbf{q}^{g}_s, \mathbf{q}^{g}_{s,gt}); \\
    \mathcal{L}_{sup} = & \sum_{i=1}^{N}\mathcal{L}_{point}+\sum_{s=1}^{N_s}\mathcal{L}_{seed},
\end{aligned}
\end{equation}
where CrossEntropy and $\mathrm{L}_2$ denote the cross-entropy and
$\mathrm{L}_2$ losses, respectively. The parameters $\lambda_g$,
$\lambda_{\mathcal{A}}$, $\lambda_w$, and $\lambda_{\mathcal{G}}$ are the
corresponding loss weights.
%
\begin{figure}[tp]
    \centering
    \includegraphics[width=\linewidth]{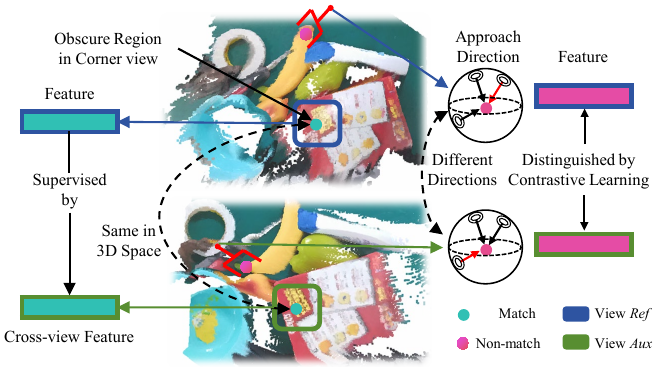}
    \caption{
    Self-supervised contrastive learning leverages the cross-view association to regularize point features.
    \textbf{Left}: For \textcolor{blue}{obscure regions} in the corner view, their features are directly supervised by the \textcolor{match}{match} points from \textcolor{green}{clear regions} in the cross views, thereby improving spatial consistency in 3D space.
    \textbf{Right}: For the same object label, the \textcolor{non-match}{non-match} point features are distinguished according to their different grasp directions, avoiding the similarity caused by supervised learning.
    }
    \vspace{-0.2in}
    \label{fig:fig3}
\end{figure}

\newline
\textbf{Self-supervised Contrastive Loss.}
The encoding of point-wise features forms the foundation of the task, which focuses only on a single view and ignores the geometric relationships in 3D space.
Building upon contrastive learning~\cite{Self-Supervised, grasp-self-supervise}, we further leverage cross-view association to formulate a self-supervised objective.
%
Specifically, we use a shared backbone to encode the point clouds from cross views, and a large number of point matches $N_{mat}$ and non-matches $N_{non}$ are generated between views, during the training time.
The point matches represent the corresponding points in 3D space, which are used to define the spatial consistency loss as:
\begin{equation}
    \mathcal{L}_{con} = \frac{1}{N_{mat}} \sum_{N_{mat}} ||f_{mat}^{ref}-f_{mat}^{aux}||_2^2 ,
\end{equation}
where the superscript denotes the source view of the point features.
We define the non-matches as point pairs on the same object with significantly different grasp approach directions to obtain the direction-discriminative loss as:
\begin{equation}
    \mathcal{L}_{dis} = \frac{1}{N_{non}} \sum_{N_{non}} max(0, M-||f_{non}^{ref}-f_{non}^{aux}||_2)^2 ,
\end{equation}
where $M = 1 - \cos(\theta)$ denotes the adaptive margin parameter used to distinguish between different approach directions, and $\theta$ is the included angle between these directions.
The total loss to train the network is as follows:
\begin{equation}
    \mathcal{L} = \mathcal{L}_{sup} + \lambda_{self} (\mathcal{L}_{con} + \mathcal{L}_{dis}),
\end{equation}
where $\lambda_{self}$ is the weight of self-supervised learning.

Compared with using only supervised learning, introducing contrastive learning provides two main advantages, as shown in Fig.~\ref{fig:fig3}.
First, for partially visible regions in corner views, the encoding is explicitly supervised by the corresponding point features from cross-view correspondences, which helps to improve the spatial consistency of point features in 3D space, preventing the corner-view features from deviating from the training distribution and thus harming generalization.
Second, for graspable points on the same object, identical supervised labels make the point features less distinctive. 
Contrastive learning mitigates this by separating features with an adaptive margin according to directional differences, which improves the subsequent prediction of grasp approach directions.
%


\section{Experiment}
\label{sec:experiment}
\begin{table*}[]
\caption{Evaluation results on GraspNet-1Billion RealSense/Kinect test data.}
\vspace{-0.1in}
\resizebox{\textwidth}{!}{%
\begin{tabular}{l|ccc|ccc|ccc}
\toprule
                                      & \multicolumn{3}{c}{\large Seen}                                                                            & \multicolumn{3}{c}{\large Similar}                                                                         & \multicolumn{3}{c}{\large Novel}                                                                           \\
                                       & \textbf{AP}                              & AP$_\textbf{0.8}$                         & AP$_\textbf{0.4}$                       & \textbf{AP}                       & AP$_\textbf{0.8}$                      & AP$_\textbf{0.4}$                  & \textbf{AP}                      & AP$_\textbf{0.8}$                  & AP$_\textbf{0.4}$                   \\
\midrule
GG-CNN                            & 15.48/16.89                     & 21.84/22.47                     & 10.25/11.23                     & 13.26/15.05                     & 18.37/19.76                     & 4.62/6.19                       & 5.52/7.38                       & 5.93/8.78                       & 1.86/1.32                       \\
Chu et al                       & 15.97/17.59                     & 23.66/24.67                     & 10.80/12.74                     & 15.41/17.36                     & 20.21/21.64                     & 7.06/8.86                       & 7.64/8.04                       & 8.69/9.37                       & 2.52/1.76                       \\
GPD                             & 22.87/24.38                     & 28.53/30.16                     & 12.84/13.46                     & 21.33/23.18                     & 27.83/28.64                     & 9.64/11.32                      & 8.24/9.58                       & 8.89/10.14                      & 2.67/3.16                       \\
PointNetGPD                 & 25.96/27.59                     & 33.01/34.21                     & 15.37/17.83                     & 22.68/24.38                     & 29.15/30.84                     & 10.76/12.83                     & 9.23/10.66                      & 9.89/11.24                      & 2.74/3.21                       \\
GraspNet                & 27.56/29.88                     & 33.43/36.19                     & 16.95/19.31                     & 26.11/27.84                     & 34.18/33.19                     & 14.23/16.62                     & 10.55/11.51                     & 11.25/12.92                     & 3.98/3.56                       \\
TransGrasp                       & 39.81/35.97                     & 47.54/41.69                     & 36.42/31.86                     & 29.32/29.71                     & 34.80/35.67                     & 25.19/24.19                     & 13.83/11.41                     & 17.11/14.42                     & 7.67/5.84                       \\
TSB(Ma)                           & 58.95/49.42                     & 68.18/58.38                     & 54.88/43.66                     & 52.97/41.49                     & 63.24/50.29                     & 46.99/34.6                      & 22.63/15.35                     & 28.53/19.18                     & 12.00/7.98                      \\
GSNet                             & 65.70/61.19                     & 76.25/71.46                     & 61.08/56.04                     & 53.75/47.39                     & 65.04/56.78                     & 45.97/40.43                     & 23.98/19.01                     & 29.93/23.73                     & 14.05/10.60                      \\
EconomicGrasp                   & 68.21/62.59                     & 79.60/73.89                     & 63.54/55.99                     & 61.19/51.73                     & 73.60/62.70                      & 53.77/43.45                     & 25.48/19.54                     & 31.46/24.24                     & 13.85/11.12                     \\
ZeroGrasp                        & 70.53/\quad-\quad\quad          & 82.28/\quad-\quad\quad          & 64.26/\quad-\quad\quad          & 62.15/\quad-\quad\quad          & 74.26/\quad-\quad\quad          & 54.97/\quad-\quad\quad          & 26.46/\quad-\quad\quad          & 33.13/\quad-\quad\quad                       & 15.11/\quad-\quad\quad    \\
\rowcolor{gray!20} 
Ours (2 views)                               & \textbf{74.08}/\textbf{64.20}                     & \textbf{85.02}/\textbf{75.51}                     & \textbf{69.34}/\textbf{56.35}                     & \textbf{62.38}/\textbf{53.41}                     & \textbf{74.92}/\textbf{64.67}                     & \textbf{55.12}/\textbf{43.84}                     & \textbf{27.27}/\textbf{21.38}                     & \textbf{33.94}/\textbf{26.64}                     & \textbf{15.15}/\textbf{11.23}                     \\
\bottomrule
\end{tabular}
}
\vspace{-0.1in}
\label{tab:table1}
\end{table*}
\begin{figure*}[]
  \centering
   \includegraphics[width=\linewidth]{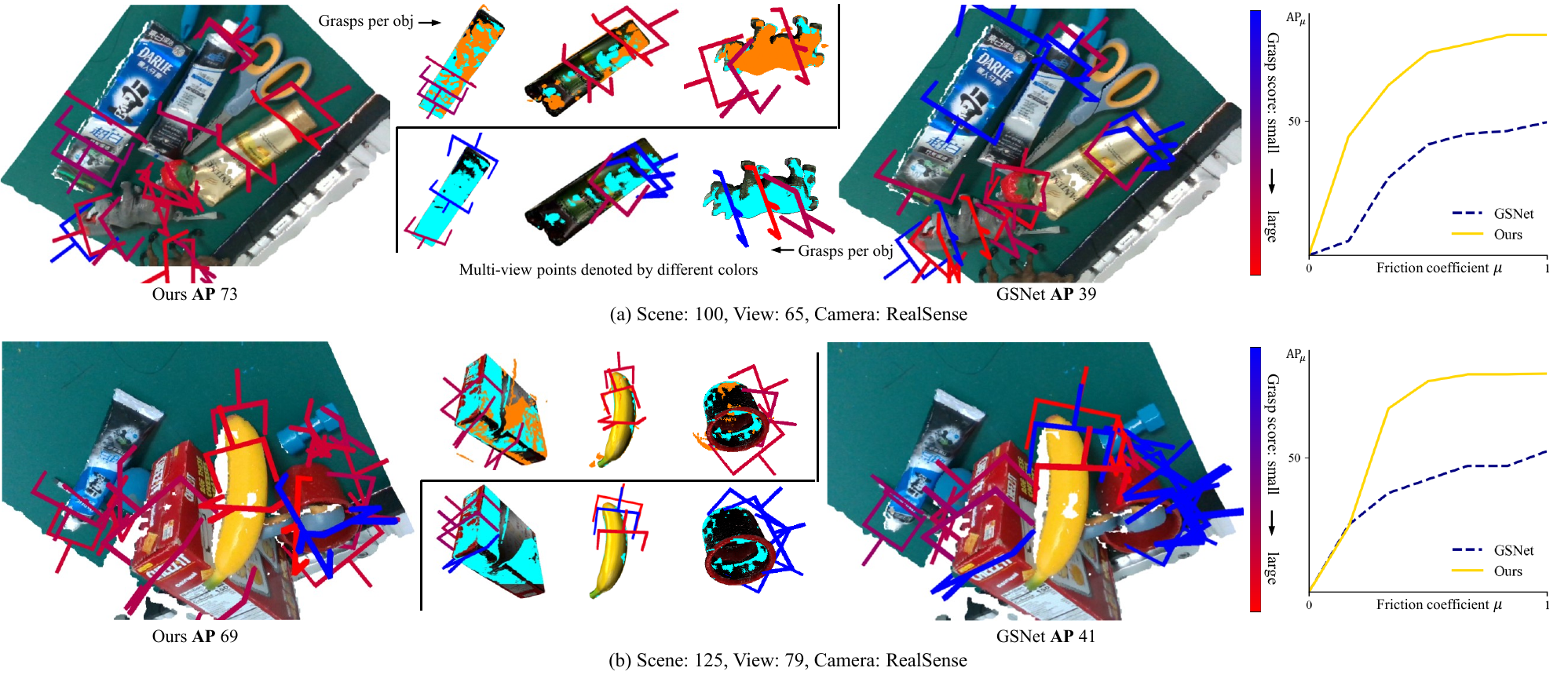}
   \vspace{-0.2in}
   \caption{
   \textbf{
   Comparison of grasp estimation results between our method and GSNet.
   }
The grasps for entire scenes and individual objects are shown for clarity.
In the scene view, our approach predicts more high-score grasps (\textcolor{MyDarkRed}{red}), while GSNet yields many low-score ones (\textcolor{MyDarkBlue}{blue}), as reflected in the overall \textbf{AP} and the AP$_\mu$ curves on the right.
In the object view, our approach combines cross-view point clouds (denoted in \textcolor{cyan}{cyan} and \textcolor{orange}{orange}), improving grasp robustness against partial observations.
}
\vspace{-0.2in}
   \label{fig:fig5}
\end{figure*}

\subsection{Implementation Details}
Given a scene point cloud obtained by back-projecting an RGB-D image, we randomly sample $N=15000$ points and extract features of dimension $C=512$. 
The number of sampled grasp seed points is $N_s=1024$, and the number of predefined approaching directions is $N_\mathcal{A}=300$, sampled uniformly on a unit sphere.
The number of neighbor points for cylinder grouping and sampling is $K=8$, 
and the numbers of in-plane rotations and approaching depths are $N_r=12$ and $N_d=4$, respectively.
The loss weight coefficients are set as $\lambda_g=10$, $\lambda_\mathcal{A}=100$, $\lambda_w=10$, $\lambda_\mathcal{G}=15$, and  $\lambda_{self}=0.2$.
%
We train the model for 8 epochs on the GraspNet-1Billion~\cite{graspnet} training set.
At the test time, the auxiliary view is randomly sampled from the 256 views available of the same scene.
All code is implemented in PyTorch and executed on one NVIDIA RTX TITAN GPU.

\subsection{Experiments on GraspNet-1Billion Benchmark}
\label{sec:datasets_evaluation}
\begin{figure}[tp]
    \centering
    \includegraphics[width=\linewidth]{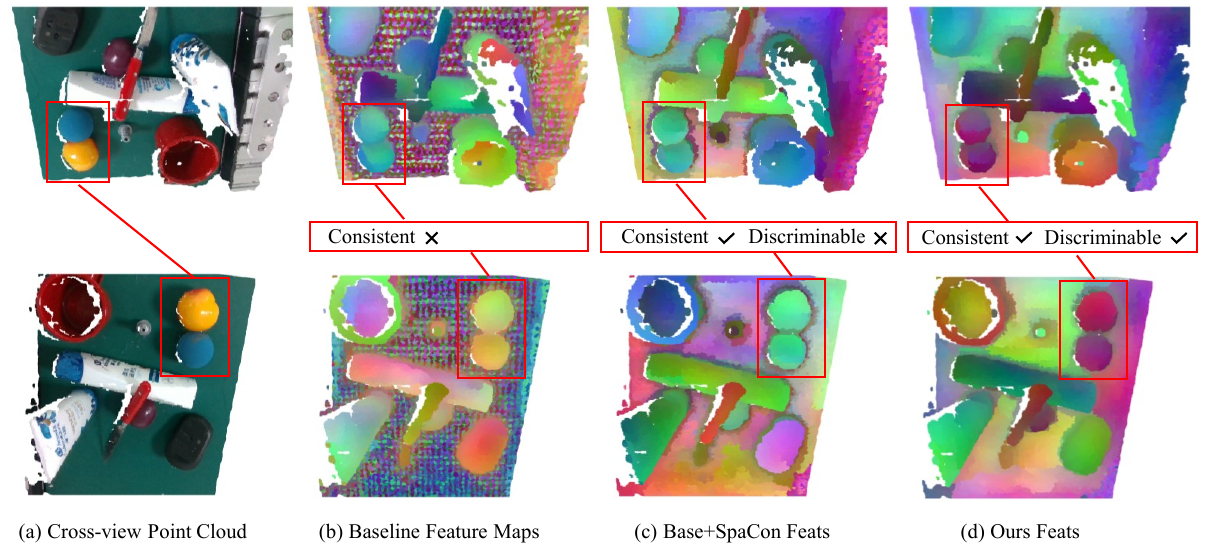}
    \caption{
Visualization of feature maps via t-SNE. 
(a) Cross-view point clouds. 
(b) Feature maps extracted by the baseline network, which fail to maintain cross-view consistency. 
(c) Feature maps learned with only the spatial consistency loss achieve cross-view consistency but lack discriminability. 
(d) Feature maps learned with the complete self-supervised contrastive learning achieve both cross-view consistency and discriminability.
}
\vspace{-0.5in}
    \label{fig:feature_map}
\end{figure}

\noindent
\textbf{Benchmark.}
GraspNet-1Billion \cite{graspnet} is a large-scale dataset for 6-DoF grasp pose evaluation, comprising 190 scenes captured from 256 viewpoints with two cameras (RealSense and Kinect). 
The first 100 scenes form the training split, while the remaining 90 are used for testing.
%
Test splits are organized by object familiarity, with each category (Seen, Similar, Novel) containing 30 scenes.
The metric AP$_\mu$ denotes the average precision given a friction coefficient $\mu$ (higher $\mu$ corresponds to easier grasps), and \textbf{AP} is the mean over all $\mu$.
\newline
\textbf{Results.}
We present the evaluation results on the Seen, Similar, and Novel splits captured by RealSense/Kinect cameras in Table~\ref{tab:table1}, achieving significant gains on all the splits.
%
%
For instance, the \textbf{AP} improves by 3.55, 0.23, and 0.81 on the Seen, Similar, and Novel splits, respectively, over the state-of-the-art results on RealSense data achieved by ZeroGrasp~\cite{grasp-reconstruction-2}.
Similarly, on Kinect data (\textit{SOTA} achieved by EconomicGrasp~\cite{grasp-reconstruction-2}), our \textbf{AP} increases by 1.61, 1.68, and 1.84 on the Seen, Similar, and Novel splits, respectively. 
The visualization of 6-DoF grasp pose estimations are presented in Fig.~\ref{fig:fig5}.
\newline
\noindent
\textbf{Ablation Study.}
\begin{table*}[]
\centering
\caption{Ablation study results on RealSense test data.}
\vspace{-0.1in}
\resizebox{\linewidth}{!}{%
\begin{tabular}{cc|ccc|c|ccc|ccc|ccc|cc}
\toprule
\multicolumn{2}{c}{ Self-supervision }  &    \multicolumn{3}{c}{ Cross-view Integration}    & \textit{Auxi-}        & \multicolumn{3}{c}{Seen}                                             & \multicolumn{3}{c}{Similar}                & \multicolumn{3}{c}{Novel} & \multicolumn{2}{c}{$\Delta$\textbf{AP} over}    \\
\small SpaCon    &\small DirDis       & \small SimAli     &  \small CylReg &  \small  AltAtt   & \textit{view}         & \textbf{AP}           & AP$_\textbf{0.8}$         & AP$_\textbf{0.4}$                       & \textbf{AP}      & AP$_\textbf{0.8}$         & AP$_\textbf{0.4}$                  & \textbf{AP}     & AP$_\textbf{0.8}$         & AP$_\textbf{0.4}$    & \textit{prev. row}   & baseline             \\
\midrule
& & &   &  &  & 63.80         &76.12       & 55.35                       & 56.90     & 70.17         & 46.40          & 23.80      & 30.15        & 11.69     & 0.00   & 0.00          \\
& & &   &  & \ding{51} & 64.34         &76.96       & 56.04                       & 57.11     & 70.61         & 46.89          & 24.03      & 30.49        & 11.87     & 0.32   & 0.32          \\
\midrule
\ding{51} &  &   &       &              &   & 67.26          &   80.07    &  60.12      &  58.69    &  72.24      &   48.81     &  25.42     &   32.07         &   13.92  & 1.97 & 2.29        \\
\ding{51} & \ding{51} &   &       &              &   & 69.36            & 81.70          & 62.43                       & 59.53       & 72.84       & 49.47          & 26.08      & 32.78           & 14.46   & 1.20 & 3.49         \\
\ding{51} & \ding{51} & \ding{51}  &       &               & \ding{51} & 70.61          &  82.14        &   63.92               &  60.21      &  73.46      &  50.22         & 26.61      &  33.14         & 14.83   & 0.82 & 4.31       \\
\ding{51} & \ding{51} & \ding{51}  &  \ding{51}     &              & \ding{51}  & 72.39            &  83.68        &   67.18           &  61.93      & 74.08       &  53.06         &  27.12     &  33.79         &  15.11    & 1.34   & 5.65    \\
\rowcolor{gray!20} 
\ding{51} & \ding{51} & \ding{51} &\ding{51}       &    \ding{51}          & \ding{51}  & \textbf{74.08}     & \textbf{85.02}         & \textbf{69.34}                    & \textbf{62.38}       & \textbf{74.92}      & \textbf{55.12}            & \textbf{27.27}      & \textbf{33.94}          & \textbf{15.15}  &   0.76 & 6.41     \\
\bottomrule
\end{tabular}
}
\small 
\textbf{Note}: (1) The upper table shows the performance of direct pre-fusion.
(2) The lower table reports the effectiveness of each component in our cross-view fusion framework, with the last two columns indicating its \textbf{AP} gain and the overall improvement of the variant.
\vspace{-0.1in}
\label{tab:table2}
\end{table*}
The significant performance improvement can be attributed to two aspects: (\textbf{i}) the self-supervised contrastive learning leverages cross-view associations to enable the point features to be spatially consistent and direction-discriminative, which is important for better generalization to corner views.
(\textbf{ii}) The cross-view-aligned cylinder integration enhances the grasp seed representation by incorporating cross-view geometric context within the grasp cylinder region, which alleviates the partial observation limitation of a single view.
%
We evaluate the effectiveness of each component in the above two aspects on the RealSense test set.
The self-supervised loss consists of the spatial consistency loss (\textit{SpaCon}) and the direction-discriminative loss (\textit{DirDis}), while the cross-view integration includes similarity-based alignment (\textit{SimAli}), cylindrical coordinate registration (\textit{CylReg}), and alternating local self-attention and seed cross-attention (\textit{AltAtt}). 
We also evaluate direct pre-fusion with an auxiliary view to further validate the effectiveness of our cross-view fusion framework.
The corresponding ablation results are presented in Table~\ref{tab:table2}.
\newline
\noindent
\textbf{Results and Analysis.}
The results show that direct pre-fusion yields only a 0.32 increase in \textbf{AP}, underscoring the significance of our framework.
We find that the limited improvement is primarily caused by the loss of geometric details during the pre-fusion process and the network’s poor generalization to cross-view relative transformations.
In contrast, our cross-view fusion framework avoids these problems and each component plays an important role in improving the performance of 6-DoF grasp estimation.
Compared with the baseline, the self-supervised contrastive learning increases \textbf{AP} by 5.56, 2.63, and 2.28 on the Seen, Similar, and Novel splits, using only single-view observations.
%
Such performance can be attributed to the regularization of point features through cross-view associations, as qualitatively illustrated in Fig.~\ref{fig:feature_map}.

By incorporating an auxiliary view using the cross-view-aligned cylinder integration module, the \textbf{AP} is further improved by 4.72, 2.85, and 1.19, respectively.
More importantly, these improvements are particularly pronounced for samples captured from challenging corner views.
 \begin{figure*}[]
  \centering
   \includegraphics[width=\linewidth]{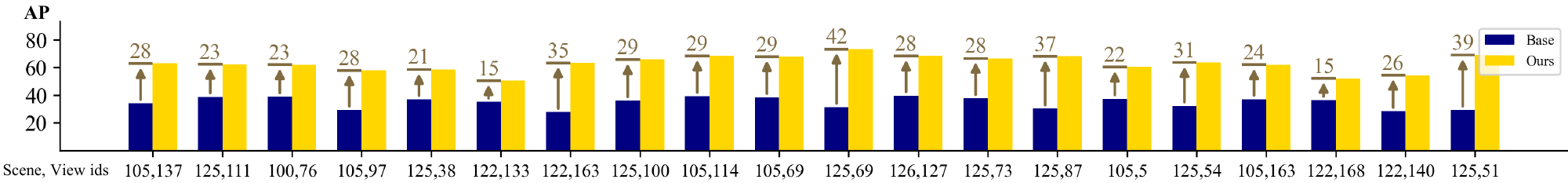}
   \vspace{-0.3in}
   \caption{
   \textbf{AP improvements in corner views.} 
   The horizontal axis represents scene-view indices where the single-view baseline yields an \textbf{AP} below 40.
   We achieve an average gain of 28.11 in these corner views, which significantly exceeds the average gain (10.28), demonstrating robustness against challenging samples by our cross-view fusion framework.
   }
   \vspace{-0.2in}
   \label{fig:fig4}
\end{figure*}
As shown in Fig.~\ref{fig:fig4}, for the 20 samples randomly selected from the Seen split where the baseline achieves an \textbf{AP} below 40 (out of 197 in total),
the improvements far exceed the average gain of 10.28 over the Seen dataset, with most gains exceeding 20 points.
Such performance validates the significant robustness of our framework by incorporating cross-view observations, capturing richer geometric context and predicting accurate 6-DoF grasp poses.
\begin{figure}[tp]
    \centering
    \includegraphics[width=0.95\linewidth]{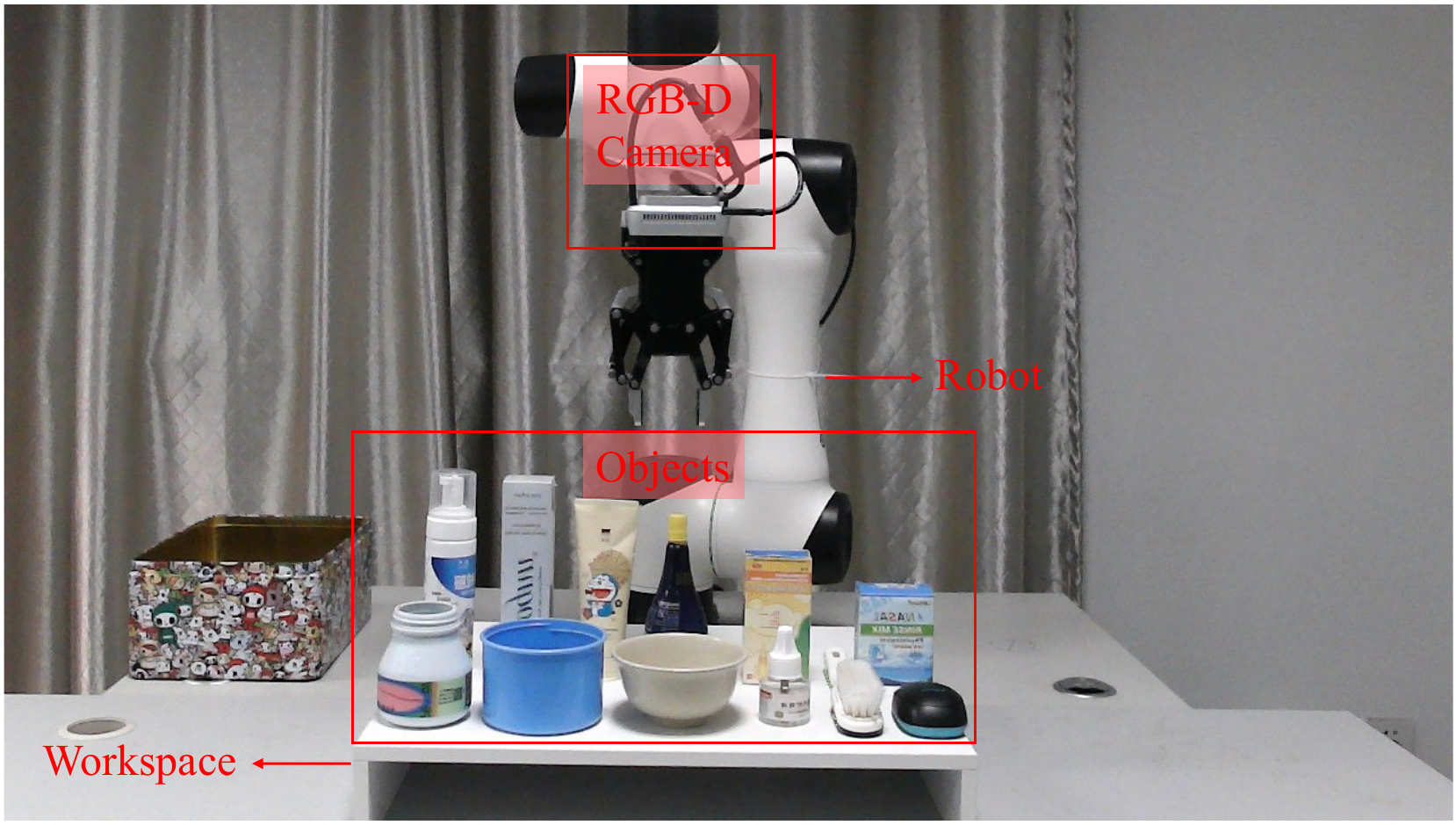}
    \vspace{-0.1in}
    \caption{
    Real-world experimental setup, including the robot, RGB-D camera, and 12 unseen grasping objects.
    }
    \vspace{-0.1in}
    \label{fig:fig6}
\end{figure}

\subsection{Experiments on Real-world Robot}
\begin{figure}[tp]
    \centering
    \includegraphics[width=0.95\linewidth]{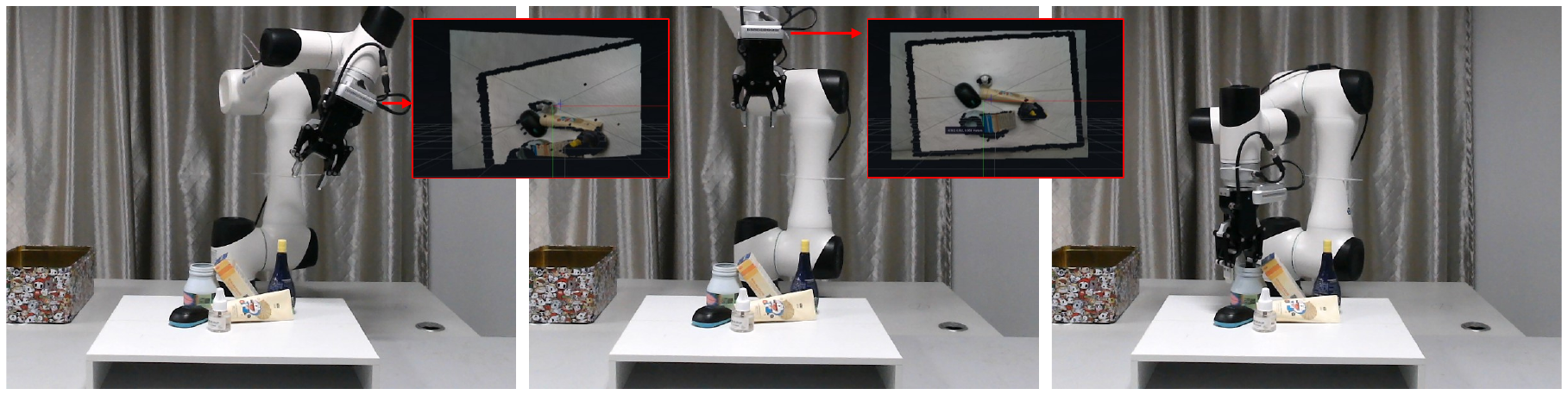}
    \vspace{-0.1in}
    \caption{
Real-world grasping pipeline showing the auxiliary (left) and primary (middle) views, and the execution (right).
Red boxes indicate the point cloud observations used for grasp estimation.
    }
    \label{fig:fig7}
    \vspace{-0.1in}
\end{figure}

The real-world experimental setup is shown in Fig.~\ref{fig:fig6}, which includes a Dobot CR5 robot, an Intel RealSense D435 camera, and 12 unseen objects of various shapes used for grasping experiments.
\noindent
We randomly select 4–6 objects to form a scene, and let the robot execute the grasp with the highest predicted score. 
After removing the grasped object from the workspace, the estimation–execution cycle is repeated until the scene is cleared.
The clutter removal experiment is a standard real-world robotic evaluation widely adopted in previous grasp estimation research.
Each method is repeated three times in every scene, and the number of grasp attempts executed by the robot is recorded to compute the grasp success rate, as presented in Table~\ref{tab:table3}.
\begin{table}[]
\caption{
Real-world clutter removal results.
}
\vspace{-0.1in}
\setlength{\tabcolsep}{2pt} 
\renewcommand{\arraystretch}{1.0} 
\centering
\resizebox{\linewidth}{!}{ 
\begin{tabular}{l|ccccc|c}
\toprule
         & \textit{S}\#1\small{(4 obj)}                      & \textit{S}\#2\small{(4 obj)}                      & \textit{S}\#3\small{(5 obj)}                      & \textit{S}\#4\small{(5 obj)}                     & \textit{S}\#5\small{(6 obj)}                     & \textit{SR} \\
\midrule
GraspNet & 4\textbackslash{}5\textbackslash{}5 & 5\textbackslash{}4\textbackslash{}4 & 6\textbackslash{}7\textbackslash{}6 & 7\textbackslash{}7\textbackslash{}8 & 8\textbackslash{}9\textbackslash{}8 & 77\%        \\
GSNet    & 4\textbackslash{}4\textbackslash{}4 & 4\textbackslash{}5\textbackslash{}4 & 8\textbackslash{}5\textbackslash{}5 & 7\textbackslash{}8\textbackslash{}6 & 7\textbackslash{}8\textbackslash{}8 & 82\%      \\
Ours     & 4\textbackslash{}4\textbackslash{}4 & 4\textbackslash{}4\textbackslash{}4 & 5\textbackslash{}5\textbackslash{}5 & 6\textbackslash{}5\textbackslash{}5 & 7\textbackslash{}6\textbackslash{}7 & 96\%     \\
\bottomrule
\end{tabular}
}
\small 
\textbf{Note}: The first row lists the scene index and object count. Table values are the number of grasp executions out of three trials. SR is the overall success rate.
\vspace{-0.2in}
\label{tab:table3}
\end{table}
Our framework demonstrates significantly higher accuracy than the baselines, achieving a 96\% success rate, which is 14\% and 19\% higher than GSNet~\cite{graspness} and GraspNet-Baseline~\cite{graspnet}, respectively.
%
For all the methods, the primary observation point is located 0.6 m above the center of the workspace.
For our framework, an auxiliary observation point is placed 0.4 m to the lower right of the center point, as shown in Fig.~\ref{fig:fig7}.
The grasping runtime is reported in Table~\ref{tab:table4}, demonstrating that our method significantly reduces reconstruction overhead (5.4s$\rightarrow$1.2s). Despite incurring an additional 1.4s compared to the single-view baseline, our approach achieves a 14\% SR improvement (Table~\ref{tab:table3}), indicating a favorable efficiency–accuracy trade-off.
We further provide qualitative demonstrations of our method in the supplementary video.
%

\begin{table}[]
\caption{The average time costs (in \textit{sec}) for real-world grasping.}
\vspace{-0.1in}
\centering
\resizebox{\columnwidth}{!}{%
\begin{tabular}{c|ccc|c}
\toprule
                 & Reconstruction & Inference & Grasp Execution & Total \\
\midrule
Single-view & -        & 0.0024         & 3.2       &   3.2     \\
\midrule
Multi-view (6 views)   & 5.4        & 0.0038         & 3.1       &  8.5         \\
\rowcolor{gray!20} 
Ours            & 1.2         & 0.0041       & 3.4       & 4.6    \\
\bottomrule
\end{tabular}
}
\vspace{-0.1in}
\label{tab:table4}
\end{table}
\noindent In particular, the relative positions of the objects are kept as consistent as possible across scene repetitions to reduce experimental variance.

\section{Conclusion}
\label{sec:conclusion}
In this paper, we propose a robust 6-DoF grasp estimation framework that combines an auxiliary-view observation to overcome the incompleteness in corner views and enhance robustness.
In contrast to the previous fusion of multi-view point clouds, our framework adopts a post-fusion pipeline that is more time-efficient for robotic grasping and preserves geometric details.
To discover the cross-view associations, we introduce a self-supervised contrastive learning strategy that regularizes the point features to be spatially consistent and direction-discriminative,
enhancing the robustness of corner views and facilitating cross-view fusion.
We further propose a cross-view-aligned cylinder integration module that fuses cross-view contextual information into a comprehensive representation for grasp prediction, comprising similarity-based alignment, cylindrical coordinate registration, and alternating local self-attention and seed cross-attention layers.
Our framework is evaluated on a large-scale and general benchmark, achieving significant performance gains, validating the contribution of each module, and demonstrating strong generalization capability in real-world robotic grasping experiments.
\section*{Acknowledgements}
This work was supported by the National Key R\&D Program of China No.~2024YFC3015801 and the National Science Fund of China under Grant Nos.~62361166670, 62276144 and U24A20330.
{\small
\bibliographystyle{IEEEtran}
\bibliography{11_references}
}

\end{document}